\documentclass{article}
  \usepackage{spconf,amsmath,graphicx}
  \usepackage{times}
  \usepackage{latexsym}
  
  \usepackage{url}

  \usepackage{bm}
  \usepackage{amsfonts}
  
  \usepackage{multirow}
  \usepackage{amsmath}
  \usepackage{epstopdf}
  \usepackage{graphicx}
  \usepackage{algorithm}
  \usepackage{algorithmic}
  \usepackage[table,xcdraw]{xcolor}
  \usepackage[normalem]{ulem}
  \useunder{\uline}{\ul}{}
  \usepackage{pdfpages}
  \graphicspath{ {img/} }
  \newcommand\BDS{{\it BDS}}
  \newcommand\Match{{\it Match}}
  \newcommand\ReRank{{\it ReRank}}
  \newcommand\Rule{{\it Rule}}
  \newcommand\Stacking{{\it Stacking}}
  \newcommand\apicall{{\tt api\_call}}
  \DeclareMathOperator*{\argmax}{argmax}
   
  \title{Context-Aware Dialog Re-Ranking for Task-Oriented Dialog Systems}
\name{Junki Ohmura$^{\star \dagger}$ \qquad Maxine Eskenazi$^{\dagger}$}

\address{$^{\star}$ Sony Corporation, Tokyo, Japan  \\
$^{\dagger}$Carnegie Mellon University, Pittsburgh, PA, USA \\
\small{\texttt{Junki.Ohmura@sony.com, max+@cs.cmu.edu}.}
}

\begin{document}
  %
  \maketitle
    
  \begin{abstract}
    Dialog response ranking is used to rank response candidates by considering their relation to the dialog history.
    Although researchers have addressed this concept for open-domain dialogs, little attention has been focused on task-oriented dialogs.
    Furthermore, no previous studies have analyzed whether response ranking can improve the performance of existing dialog systems in real human--computer dialogs with speech recognition errors.
    In this paper, we propose a context-aware dialog response re-ranking system.
    Our system reranks responses in two steps:
    (1) it calculates matching scores for each candidate response and the current dialog context;
    (2) it combines the matching scores and a probability distribution of the candidates from an existing dialog system for response re-ranking.
    By using neural word embedding-based models and handcrafted or logistic regression-based ensemble models,
    we have improved the performance of a recently proposed end-to-end task-oriented dialog system on real dialogs with speech recognition errors.
  \end{abstract}
  


   \section{Introduction}
   Spoken dialog systems must be robust enough to deal with automatic speech recognition (ASR) errors.
   Word error rate (WER) of the ASRs in dialog systems is not insignificant; e.g., around 25\%~\cite{henderson2014second}.
   Even the latest neural-based ASR models have a word error rate around 5\%~\cite{xiong2017microsoft,46687}.
   These errors make it difficult for dialog systems to understand users' intentions.
   Recently, task-oriented end-to-end dialog systems have showed good results on real human--computer dialogs~\cite{DBLP:journals/corr/BordesW16,DBLP:conf/acl/WilliamsAZ17}, however, their system employed speech transcriptions for the user utterances.
   Also, the dataset~\cite{DBLP:journals/corr/BordesW16} they used was generated from dialogs with high word error rates (around 25\%)~\cite{henderson2014second}.
   To measure robustness against ASR errors in dialog systems, various studies reported that they improved system predictions by considering; acoustic, syntactic, and semantic features~\cite{chotimongkol2001n}, ASR N-best hypotheses~\cite{yaman2007discriminative}, and word confusion networks~\cite{tur2013semantic}. 
   
   In task-oriented dialogs, dialog systems must understand the dialog history in order to serve the user's goal.
   To do this, a system extracts user preferences, stores them in memory, and uses them in the system response.
   Various dialog systems have been in this vein~\cite{raux2005let,young2006using,W16-3601,DBLP:journals/corr/BordesW16,I17-1074}.
   Yet these systems usually do not have a feedback loop to determine if the estimated system action is valid given the current dialog context.

   \begin{figure}[t!]
     \begin{center}
     \includegraphics[scale=0.18]{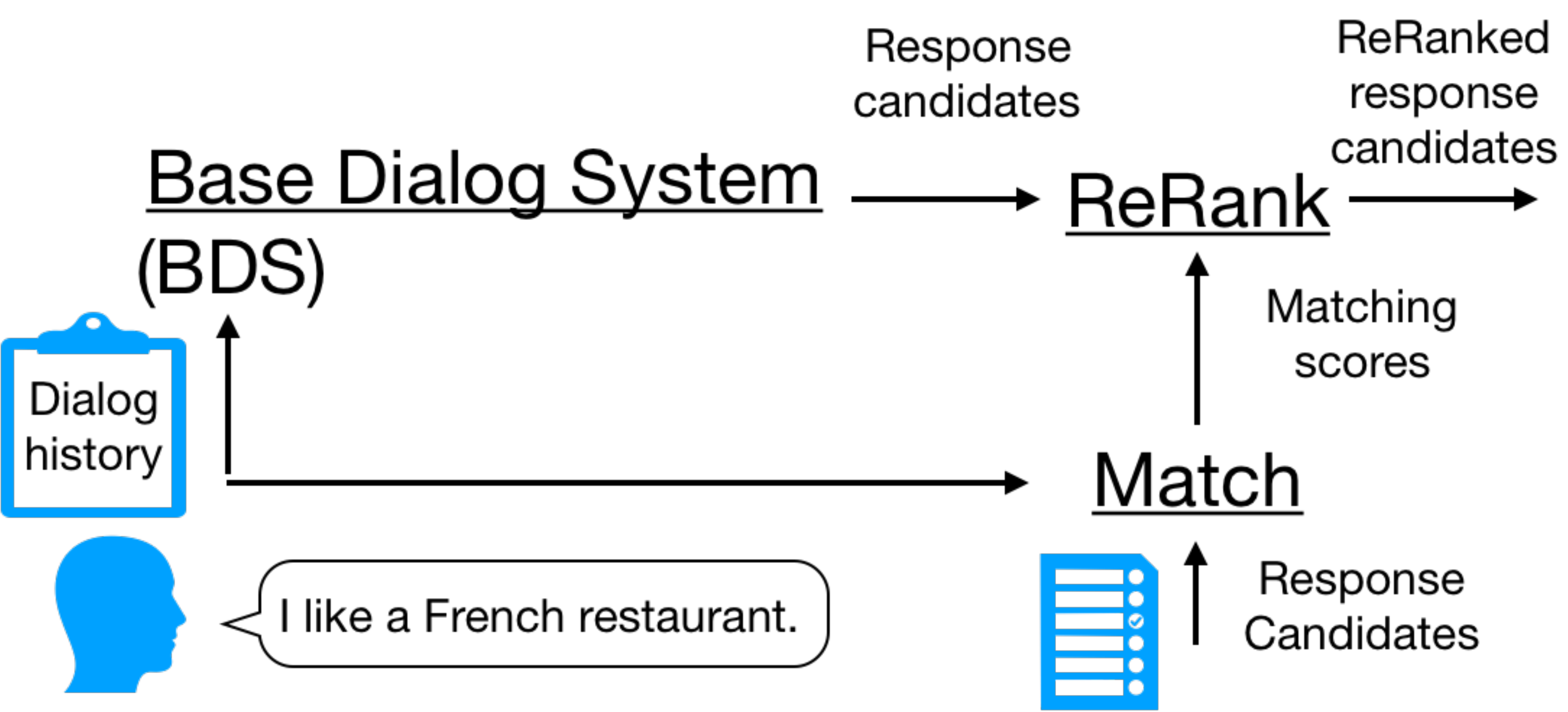}
     \end{center}
     \caption{Overview of the dialog context-aware re-ranking model}
     \label{fig:arch}
   \end{figure}
   
   Re-ranking effectively improves prediction reliability at various levels, i.e., speech recognition, domain selection, dialog state tracking, and response selection.
   Hypothesis re-ranking techniques to improve speech recognition accuracy in dialog systems have been proposed~\cite{chotimongkol2001n,jonson2006dialogue}, in which syntactic, semantic, and acoustic features for re-ranking with classifiers, such as linear regression models, are combined.
   ~\cite{nakano2011two,ryu2012hierarchical} proposed re-considering procedures for domain selection.
   They implemented an additional module for final dialog domain selection.
   Ensemble learning techniques have been used to predict dialog states~\cite{henderson2014second}, with multiple dialog state trackers combined to improve prediction. 
   Unlike previous research, our model focuses on predicting a system response and works as an additional module that does not require additional training within an existing dialog system .
   Only a few studies have been conducted on response selection tasks for task-oriented dialogs.
   For example, response ranking in task-oriented dialogs was implemented in~\cite{DBLP:journals/corr/BordesW16}; however, their ranking method was treated as a 1-best response selection task.
   For comparison, we applied their approach, including baselines, for our reranking.
   The result is described in Section \ref{seq:experiments}. 
   Response filtering was also implemented in~\cite{DBLP:conf/acl/WilliamsAZ17} by integrating the developer's hand-crafted rules into end-to-end networks. 
   
   To obtain consistent behavior, a model that uses response candidates directly as input features is also useful.
   The task of choosing a correct response among the candidate responses is called response ranking.
   Generally, there are two types of approaches:
   query-response pair base~\cite{NIPS2013_5019,DBLP:journals/corr/TanXZ15,Avoiding}
   and context-response pair base~\cite{W15-4640,DBLP:journals/corr/KadlecSK15,D16-1036,C16-1063,P17-1046}.
   The former matches a user query with candidates without using the dialog history and handles task-oriented dialogs. 
   The latter uses a dialog history-candidate pair and addresses open-domain dialogs.
   However, these systems have not been used for response re-ranking.

   This paper proposes a context-aware dialog re-ranking system that employs two additional models: \Match~and \ReRank.
   \Match~considers the response candidates to be related to the dialog context.
   \ReRank~combines different outputs, i.e., predictions from a dialog system and matching scores from \Match~for better prediction.
   We employ an existing dialog system into our framework.

   
   \section{Response Re-ranking Model}
   In this section, we describe the construction of the response re-ranking system shown in Fig.~\ref{fig:arch}.
   Our re-ranking system consists of three main modules: the base dialog system ({\it BDS}), \Match, and \ReRank.

   \subsection{Base Dialog System}
   \BDS~(Base Dialog System) is an arbitrary dialog system which predicts a system response with a probability distribution.
   In our re-ranking model, BDS is a fixed model, i.e., it does not require training for re-ranking as we assume that existing dialog systems are reused as \BDS.
   This reusability allows use of various existing models without additional training for the \BDS.
   In this work, we used end-to-end memory memory networks \cite{NIPS2015_5846, DBLP:journals/corr/BordesW16} as our \BDS.

   {\bf Memory Networks}:
   \label{seq:memnn}
   Memory Networks are neural networks that read and store events for solving tasks in the areas of natural language processing.
   Let \(\{x_1,..., x_T\}\), \(q\), and \(\{a_1,..., a_N\}\) represent the words of the input dialog history, a query, and candidates for system action, respectively, where \(T\) is the number of turns of the dialog history, and \(N\) the number of all possible system actions.
   Each variable constitutes one sentence.
   Further, \(x_i\) is embedded in a \(d\)-dimensional vector using matrix \(A\in \mathbb{R}^{d\times V}\), (where \(V\) is the vocabulary size) and is encoded into a memory vector \(m_i\in \mathbb{R}^{d}\) using {\it position encoding}, a technique reported in~\cite{NIPS2015_5846} where \(m_i\) is affected by word order.
   The query \(q\) is also embedded into query vector \(u\) using embedding \(B\in \mathbb{R}^{d\times V}\). 
   With the memory vectors and \(u\), an attention score for each item of memory \(m_i\) is given by
   \begin{equation}
     p_i = {\rm Softmax}(u^{\mathrm{T}}m_i),\ (i=1,\ldots,T).
     \label{eq:pi}
   \end{equation}
   We also have a vector \(c_i\) from \(x_i\) using matrix \(C\in \mathbb{R}^{d\times V}\). The memories are read by taking their weighted sum:
   \begin{equation}
     o = \sum_{i}p_ic_i.
   \end{equation}
   We employ \(K\) hop operations. Thus, the input to the \(k+1\)-th layer is updated by the following equation:
   \begin{equation}
     u^{k+1} = u^k + o^k.
   \end{equation}

   We also apply {\it adjacent weight tying}, e.g., \(A^{k+1}=C^k\) and \(B=A^1\). 
   Finally, the model predicts the system actions \(\hat a\) using the weight matrix \(W\), as follows:
   \begin{equation}
     \hat a=\argmax_{i}({\rm Softmax}(Wu^{K+1})).
     \label{eq:hat_act}
   \end{equation}

   \subsection{Match}
   \Match~calculates matching scores between the dialog history and the candidates generated by \BDS.
   While \BDS~predicts a system response given the dialog history, \Match~uses the candidate responses directly for the matching scores.
   Generally, these matching methods are used as response selection models such as \cite{NIPS2013_5019,DBLP:journals/corr/TanXZ15,Avoiding,W15-4640,DBLP:journals/corr/KadlecSK15,D16-1036,C16-1063,P17-1046}. 
   In our re-ranking task, we use the scores as input features of re-ranking.

   To evaluate the effect of \Match, we prepared five models which include two non-machine learning models.
   The first three models are identical to those described in \cite{DBLP:journals/corr/BordesW16}.
   The fourth model is based on Memory Networks for predicting matching scores.
   The last model, QA-LSTM, is an response selection model developed by \cite{DBLP:journals/corr/TanXZ15}.
   We now describe each of \Match~in detail.

   {\bf TF-IDF}:
   This model uses bag-of-words features to represent inputs and targets:
   the whole dialog history including the last utterance and the candidate responses, respectively.
   The matching score is the TF-IDF weighted cosine similarity between the inputs and the targets.
  
   {\bf Nearest Neighbor (NN)}:
   In this model, we consider (last utterance - response) pairs for the scoring method.
   By considering the pairs, this model attempts to find the most similar conversation in the training set.
   Word overlap is used as the scoring mechanism.
   The pairs are sorted by decreasing the co-occurrence frequency when multiple responses are linked to the same utterance in the training set.

   {\bf Supervised Embedding (SLEmb)}:
   This is a supervised word embedding method. Let \(x\) represent the words of the input dialog history including the last utterance, 
   \(y\) represents the candidate response to the input.
   Then, the scoring function is given by: \(f(x, y)=(Ax)^{\mathrm{T}}(By)\),
  where \(A\) and \(B\) are \(d\times V\) word embedding matrices, (where \(d\) is embedding size and \(V\) size of the vocabulary).
  The embedding model is trained with a margin ranking loss with negative samples.

   {\bf Match Memory Networks (MMNs)}:
   We also used Memory Networks for \Match, referred to as {\it Match Memory Networks (MMNs)} to distinguish them from \BDS.
   We modified two equations of the original Memory Networks model described in \ref{seq:memnn}.
   We first modified Equation \ref{eq:pi} by taking the L2 norm for attention, as our preliminary analysis showed that attention is biased to one or two dialog turns without the normalization.
   The second modification was the last layer (Eq.~\ref{eq:hat_act}). 
   While the original memory networks predict system actions using a weight matrix,
   MMNs calculate a cosine similarity between the dialog context and the candidate response \(a_j\).
   \(a_j\) is embedded into vector \(v_j\) via matrix \(A^{K+1}\) in the same manner as the \(q\) embedding. Thus,
   \begin{equation}
     match_j = \cos(u^{K+1}, v_j),\ \ \ (1\leq j \leq N),
     \label{eq:cos}
   \end{equation}
   where \(match_j\) represents the extent to which a given response candidate is related to the given dialog context.

\begin{figure}[ht]
    \begin{center}
    \includegraphics[scale=0.15]{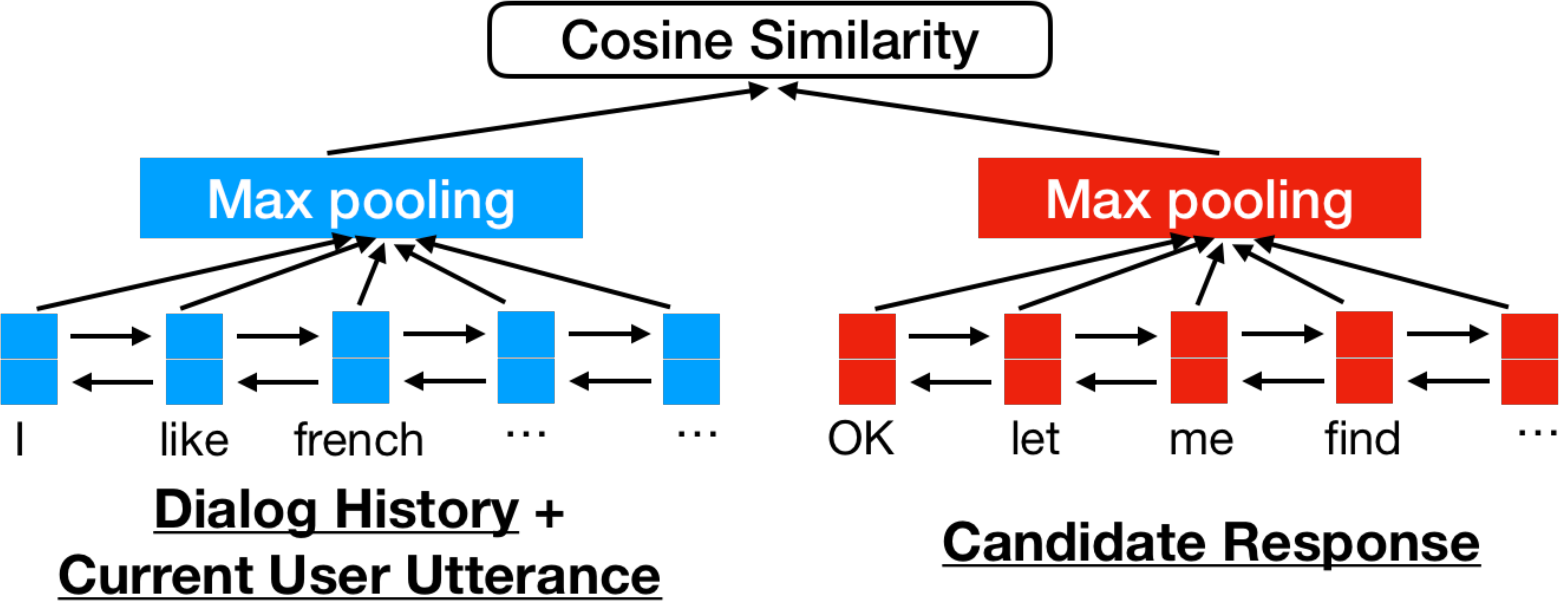}
    \end{center}
    \caption{QA-LSTM}
    \label{fig:qalstm}
  \end{figure}

   {\bf QA-LSTM}:
   QA-LSTM \cite{DBLP:journals/corr/TanXZ15} is a simple and strong model for response selection tasks (Fig.~\ref{fig:qalstm}).
   The dialog history and the candidate responses are encoded into the same word representations as Memory Networks and Supervised Embedding.
   Each input and target is fed to bidirectional long short-term memory (BiLSTM), which can generate word-level representations.
   Each output is aggregated in one of three simple ways: (1) average pooling; (2) max pooling; (3) concatenation of the last vectors of both directions.
   In this study, we used a max pooling method to aggregate each word representation.
   Finally, the cosine similarity is calculated between both aggregated representations.
   For the loss function, we used the same margin ranking loss function as for Supervised Embedding.
   The margin loss is calculated for each dialog turn:
   \begin{equation}
    \mathcal{L}_{match}=\max (0, m - match_+ + match_-),
   \end{equation}
   where \(m\) represents a constant margin (a hyperparameter) between the scores of correct and incorrect system action pairs. \(match_+\) and \(match_-\) return a matching score for correct and incorrect action candidates, respectively, using the same dialog context given in Eq.~\ref{eq:cos}. 

	Regarding the margin ranking in MMNs and QA-LSTM, a correct response is obtained from a ground truth response. An incorrect response is randomly sampled from all responses in the dataset. The incorrect response is sampled until the loss \(L > 0\) or the number of samplings reaches a specified value.

   \subsection{ReRank}
   \ReRank~reranks the candidate responses by considering the two predictions from \BDS~and \Match.
   To combine the two outputs, we developed two \ReRank~models: a rule-based and a logistic regression model.

   {\bf Rule}:
   \Rule~is a heuristically-designed method for combining the two different \BDS~and \Match~outputs.
   This model calculates \(score_i\) as a re-ranked score:
   \begin{equation} 
     score_i={\rm Normalize}(p_i)\times (\alpha_i\times matching\_score_i),
     \label{eq:rule-score}
   \end{equation}
   where \(p_i\) is the probability of \BDS~candidate \(a_i\), and \(\alpha_i\) is a scalar reflecting of the rank of \(a_i\) of the matching score of \Match.
   The highest scoring response is selected as the final response.

\begin{figure}[ht]
    \begin{center}
    \includegraphics[scale=0.16]{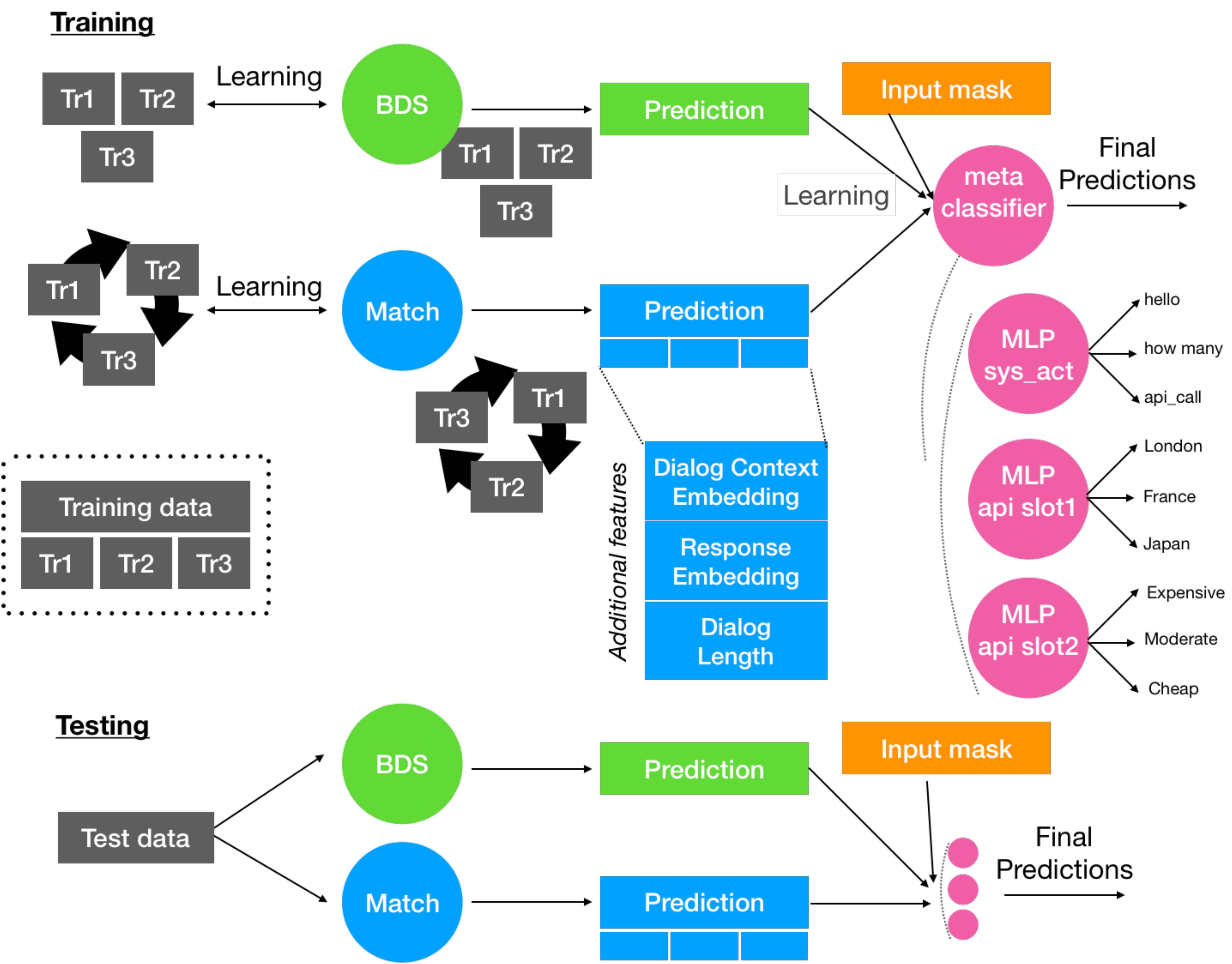}
    \end{center}
    \caption{Stacking and logistic regression}
    \label{fig:stacking}
  \end{figure}

   {\bf Stacking and logistic regression}:
   We tackled this problem as ensemble learning, a technique that uses multiple learning algorithms to obtain better prediction.
   Among the various techniques, we used \Stacking~\cite{wolpert1992stacked} to combine two models.
   \Stacking~gives the predictions of multiple models given as input to a second-level learning model.
   Our \Stacking~ model has two learning layers: (1) training of base-level classifiers; (2) training of meta-classifiers (Fig. \ref{fig:stacking}).
   We can use this meta-classifier as our \ReRank.
   We explain this model with Fig.~\ref{fig:stacking}.
   
   Both \BDS~and \Match~are treated as base classifiers in \Stacking~system.
   To train the models, the training data are split into folds (subsets).
   \Match~is trained on training subset 1 and 2. The trained \Match~output is the prediction for training subset 3.
   This process is repeated until all subsets are used for the predictions.
   In our experiments, \BDS~was trained on the entire training set at once since we took this model as a fixed model. 
   Note that it is not necessary for \BDS~to be trained in the first stacking layer; i.e., our system can afford to reuse existing models trained on different datasets.
   After training the base classifiers, we obtain two predictions from \BDS~and \Match; the former is a probability distribution over system responses \(y_{bds}\), and the latter is matching scores \(y_{mat}\) between the dialog history and the candidate responses.

   The meta classifiers use the predictions from the base classifiers for the final prediction.
   We apply multiple logistic regression (LR) models as the classifier.
   There are many similar \apicall~ system actions in the bAbI dialog dataset (Section~\ref{seq:dataset}) \cite{DBLP:journals/corr/BordesW16}, which are different from slot entities.
   Note that {\tt api\_call} is a special system action for searching restaurant information, taking multiple slots as its arguments.
   To capture similarity in the meta classifiers, we developed multiple LR models and simplified the {\tt api\_call} actions as one action.
   The first LR predicts the simplified system actions.
   The remaining LRs predict each slot in {\tt api\_call}, e.g., cuisine type, location, and price.
   The output from the slot LRs is used to reconstruct the original {\tt api\_call} actions, where the first LR predicting the current response is the {\tt api\_call} with the highest score.
   In LR training, the cost function is the sum of the cross entropy of each LR.
   
   
   We found that using multiple LRs for system actions is effective for predicting similar {\tt api\_call} actions.
   In the dataset employed in this work, some {\tt api\_call} actions in the test data do not appear in the training data.
   On the other hand, all slots appear in the training data.
   Therefore, separately handling arguments works well.
   
   We also use ranking masks \(m_{bds}\) and \(m_{mat}\) (for \BDS~and \Match, respectively) to obtain the predictions, to focus on the high-score candidates.
   The mask values are set to 1.0 if the candidate is within the top \(H\) predictions; otherwise, 0.
   The masks are used for \BDS~and \Match~separately.

   We use additional features for the LR input: {\it dialog context embedding} \(e_{ctx}\), 
   {\it answer embedding}  \(e_{ans}\), and {\it the length of the dialog history} \(l\) for QA-LSTM and MMNs.
   Here, \(e_{ctx}\) is a dialog history embedding vector. 
   In MMNs, this is relevant to \(u^{K+1}\) in Eq.~\ref{eq:cos}.
   In QA-LSTM, this vector is aggregated from vectors from the BiLSTM of the dialog history.
   Similarly, \(e_{ans}\) is also an embedding vector; however, it differs in that it embeds the candidate response. 
   We choose the response embedding with the highest matching score from the \Match~output.
   Finally, \(l\) is a one-hot vector which represents the number of turns of the current dialog history.
   Then, the input of the meta classifier is:

  \begin{equation}
   input=[y_{bds}\odot m_{bds}, y_{mat}\odot m_{mat}, e_{ctx}+e_{ans}, l],
   \label{eq:stacking_input}
  \end{equation}
  where \(\odot\) denotes element-wise multiplication, \(+\) the element-wise plus, and \([\cdot, \cdot]\) the concatenation.
  The input is shared for all LRs.

  We also applied {\it majority voting} for both \ReRank~models when the highest-scoring candidates of \BDS~and \ReRank~are the same.
   

    \section{Experiments}
    \label{seq:experiments}
    In this section, we report experiments to determine whether our re-ranker improves the performance of a recently proposed end-to-end dialog system~\cite{DBLP:journals/corr/BordesW16} by considering the relation between the dialog context and the response candidates generated by the dialog system.

    \subsection{Dataset}
    \label{seq:dataset}
    We used the following datasets to assess our system:
  
    \textbf{bAbI dialog dataset}: 
    The bAbI dialog dataset~\cite{DBLP:journals/corr/BordesW16} is a set of six tasks from dialogs about restaurant reservations.
    Although Tasks 1--5 were systematically generated using the same dialog dataset, they were different in that each required different dialog skills. 
    In contrast, Task 6 is a human--computer restaurant reservation dialog dataset created by converting the Second Dialog State Tracking Challenge dataset~\cite{henderson2014second} into the bAbI dataset format.
    Thus, Task 6 is based on a real dialog dataset and incorporates speech transcription.
    We found that the speech recognition error rate was high,  as shown on Table \ref{tb:wer}.
    Therefore, to use real dialog data containing the ASR errors, 
    we reproduced a Task 6 generator by following the configurations reported in~\cite{DBLP:journals/corr/BordesW16}.
    We chose a 1-best hypothesis of the ASR results;  we call the resulting Task 6 ``ASR-Task 6''.
    \begin{table}[ht]
      \centering
      \small
      \caption{Speech recognition error analysis of DSTC2. (WER: word error rate).
      Although both recognizers are based on the same model, Recog 0 had artificially degraded acoustic models.
      Note that ``filtered'' indicates the evaluations in which stop-words were removed from the text before the evaluation.
      }
      \begin{tabular}{l|r|r|r}
      \hline
      \multicolumn{1}{c|}{} & \multicolumn{1}{c|}{\textbf{Average}} & \multicolumn{1}{c|}{\textbf{Recog 0}} & \multicolumn{1}{c}{\textbf{Recog 1}} \\ \hline
      WER                    & 25.57\%                               & 31.82\%                                    & 19.24\%                                    \\ \hline
      WER (filtered)         & 24.17\%                               & 30.03\%                                    & 18.58\%                                    \\ \hline
      \end{tabular}
      \label{tb:wer}
    \end{table}
    
    \textbf{bAbI+}: 
    bAbI+~\cite{Shalyminov.Eshghi.Lemon_2017} was created systematically by adding natural disfluencies, such as self-corrections, hesitations, and restarts to the bAbI dialog dataset.
    bAbI+ was limited to Task 1 as it focused on the capability of the system to ask users about their restaurant preferences.

    \subsection{Setup}
    We conducted two experiments on {\bf bAbI/bAbI+} involving {\bf ASR-Task 6}.
    
    \noindent
    {\bf bAbI/bAbI+}:
    In this experiment, our re-ranker was trained on the bAbI dialog Task 1 and tested on the bAbI+ dataset.
    We used the same experiment settings as in~\cite{Shalyminov.Eshghi.Lemon_2017}.
    It is known that Memory Networks can perfectly answer questions for Task 1 \cite{DBLP:journals/corr/BordesW16}.
    However, the Task 1 accuracy decreases dramatically if the models are tested with bAbI+ \cite{Shalyminov.Eshghi.Lemon_2017}.
    Furthermore, we are interested in how our models are robust to the disfluencies.
    That is why we used bAbI+ for the testing.
  
    {\bf ASR-Task 6}:
    Our re-ranker was trained on the newly created ASR-Task 6.
    
    We used training/validation/test sets that were already split by the dataset providers on both bAbI+ and DSTC2.
    
    \subsection{Implementation details}
    \noindent
    \BDS:
    We use Memory Networks as \BDS.
    In the bAbI/bAbI+ task, we employed their implementation\footnote{See https://github.com/ishalyminov/memn2n}.
    In the ASR-Task 6 task, we reproduced Memory Networks by following the configurations reported in~\cite{DBLP:journals/corr/BordesW16}.
    Basically, both models are the same.

    \Match:
    We followed the configurations reported in \cite{DBLP:journals/corr/BordesW16} for TF-IDF, Nearest Neighbor, and Supervised Embedding.
    For QA-LSTM, we used a bidirectional gated recurrent unit (GRU) \cite{cho2014learning} to obtain vectors of the dialog context and the candidate response. We used 64 GRU dimensions and shared the GRU for encoding of the dialog context and of the candidate.
    For MMNs, we used a word embedding size of 128 and set the number of hops (\(K\)) to three.
    For QA-LSTM and MMNs, the following parameters were common:
    a word embedding size of 128, a margin \(M\) of 0.5, and a training set split into five folds.
    Negative sampling was performed 100 times, on the condition that the loss \(L\) was positive for every batch.
    These models were trained with the Adam optimizer~\cite{DBLP:journals/corr/KingmaB14} for 20 epochs in each fold and a batch size of 32.
    For brevity, the \Match~was shared for both \ReRank~models; however, the rule-based \ReRank~does not necessarily require training on fold data.

    \ReRank: 
    For \Rule, to calculate \(score_i\), we used a sigmoid function to normalize the \BDS~probabilities.
    The probabilities of most candidates were almost zero, because the base model used softmax over thousands of candidates.
    As the probability of a correct response have been close to zero, we were required to change the range of the probabilities.~\footnote{The new value ranged from 0.5 to 0.731, as the probability was between 0 and 1.}
    \(\alpha\) was set to 1.0 if the score of a given candidate was within the top five, as arranged by \Match; otherwise, it was set to 0.
    For \Stacking, all the meta classifiers with the hidden dimension of 700 (one-layer perceptron) were used, with a \(H\) of 10 for the input masks.
    The training batch size was 64 for 20 epochs with the Adam.

    \subsection{Main results}
    \label{sec:res-diso}
    The accuracy results for the dialog turns are presented in Table~\ref{tb:result-acc}.
    Here, the accuracy corresponds to the ratio of correct response selection for the entire dataset.
    ``MAT'' represents the accuracy based on the score of \Match, 
    ``RR1'' corresponds to rule-based \ReRank, and ``RR2'' is for the stacking \ReRank.

    TF-IDF and NN yielded poor results for both datasets.
    It is difficult for TF-IDF to handle dialog features, such as dialog flow, as bag-of-words does not consider word order.
    The results shows that choosing a correct response without considering the current dialog context is almost impossible.
    NN uses (last utterance -- candidate response) pairs to choose the response.
    bAbI/bAbI+ has synthesized simple dialogs; therefore, the performance is better than that of TF-IDF.
    However, NN is not effective for ASR-Task 6 since it is quite rare for exactly the same pair to be found in the training dialog.

    
    SLEmb obtained almost the highest accuracy on the bAbI/bAbI+ while its accuracy on ASR-Task 6 dropped dramatically.
    It can be presumed that this change was due to the difference in vocabulary size.
    The vocabulary size of ASR-Task 6 is 1490, whereas that of bAbI/bAbI+ is 111.
    Moreover, the (context -- response) pairs were not simple since ASR-Task 6 is a corpus of real dialogs. 
    Therefore, SLEmb only works for limited vocabulary and dialog patterns, e.g., synthetically generated dialogs.

    Unlike the previous models, MMNs and QA-LSTM sufficiently improved the accuracy of \BDS~on both datasets.
    
    All MMN models outperformed \BDS.
    Unlike the previous models, Memory Networks have the ability to read dialog context.
    Further, the cosine similarity between the dialog context and the candidate responses was effective in improving predictions.

    QA-LSTM can also understand the dialog context from the  Memory Networks, i.e., the former uses recurrent neural networks, the latter uses memory reading and writing components.
    For QA-LSTM, RR1 and RR2 had the highest total accuracy scores on both datasets.
    Both the RR1 and RR2 models could combine different outputs to achieve better predictions.

    Overall, for both MMNs and QA-LSTM, the accuracy of {\tt api\_call} was dramatically boosted.
    We presume that word-wise embedding is effective since the arguments (slots) in {\tt api\_call} appear in the dialog history.
    We analyzed the effect of word-wise embedding in Section \ref{seq:analysis}.

\begin{table}[ht]
  \centering
  \small
  \caption{Dialog turn accuracy. ``Total'' represents the accuracy for an entire test dataset and ``API'' indicates the accuracy of the API call action that the dialog system decided to take.
  The best-performing models are formatted in bold, while underscores indicate the best score for each model and dataset combination.
  A yellow background indicates a result superior to that of \BDS.
  }
  \label{tb:result-acc}
  \begin{tabular}{cc|ll|ll}
  \multicolumn{1}{l}{}      & \multicolumn{1}{l|}{} & \multicolumn{2}{l|}{bAbI/bAbI+}                                                          & \multicolumn{2}{l}{ASR-Task-6}                                                      \\
  \multicolumn{1}{l}{}      & \multicolumn{1}{l|}{} & Total                                       & API                                         & Total                                       & API                                   \\ \hline
  \multicolumn{2}{c|}{BDS}                          & \textit{82.1}                               & \textit{21.7}                               & \textit{42.5}                               & \textit{35.8}                         \\ \hline
                            & MAT                   & 6.0                                         & 2.3                                         & 3.8                                         & 0.0                                   \\
                            & RR1                   & 8.2                                         & {\ul 15.6}                                  & 4.3                                         & 0.4                                   \\
  \multirow{-3}{*}{TF-IDF}  & RR2                   & {\ul 51.6}                                  & 0.4                                         & {\ul 38.6}                                  & {\ul 30.2}                            \\ \hline
                            & MAT                   & 41.8                                        & 0.0                                         & 0.3                                         & 0.0                                   \\
                            & RR1                   & 47.1                                        & 0.0                                         & 10.4                                        & 0.0                                   \\
  \multirow{-3}{*}{NN}      & RR2                   & {\ul 56.7}                                  & 0.0                                         & {\ul 38.0}                                  & {\ul 27.9}                            \\ \hline
                            & MAT                   & \cellcolor[HTML]{FFFC9E}{\ul 91.3} & \cellcolor[HTML]{FFFC9E}{\ul 51.6}          & 25.9                                        & 12.7                                  \\
                            & RR1                   & \cellcolor[HTML]{FFFC9E}90.7                & \cellcolor[HTML]{FFFC9E}47.9                & {\ul 33.5}                                  & {\ul 27.2}                            \\
  \multirow{-3}{*}{SLEmb}   & RR2                   & \cellcolor[HTML]{FFFC9E}82.3                & \cellcolor[HTML]{FFFC9E}17.6                & 25.0                                        & 5.4                                   \\ \hline
                            & MAT                   & \cellcolor[HTML]{FFFC9E}84.7                & \cellcolor[HTML]{FFFC9E}38.1                & \cellcolor[HTML]{FFFC9E}43.7                & \cellcolor[HTML]{FFFC9E}50.6          \\
                            & RR1                   & \cellcolor[HTML]{FFFC9E}{\ul 86.3}          & \cellcolor[HTML]{FFFC9E}{\ul 38.3}          & \cellcolor[HTML]{FFFC9E}{\ul 45.6}          & \cellcolor[HTML]{FFFC9E}{\ul 53.9}    \\
  \multirow{-3}{*}{MMNs}    & RR2                   & \cellcolor[HTML]{FFFC9E}86.0                & \cellcolor[HTML]{FFFC9E}36.6                & \cellcolor[HTML]{FFFC9E}44.8                & \cellcolor[HTML]{FFFC9E}43.2          \\ \hline
                            & MAT                   & \cellcolor[HTML]{FFFC9E}92.0                & \cellcolor[HTML]{FFFC9E}{\ul \textbf{60.9}} & \cellcolor[HTML]{FFFC9E}46.2                & \cellcolor[HTML]{FFFC9E}{\ul \bf 70.3}          \\
                            & RR1                   & \cellcolor[HTML]{FFFC9E}88.8                & \cellcolor[HTML]{FFFC9E}49.6                & \cellcolor[HTML]{FFFC9E}{\ul \textbf{48.7}} & \cellcolor[HTML]{FFFC9E}68.1 \\
  \multirow{-3}{*}{QA-LSTM} & RR2                   & \cellcolor[HTML]{FFFC9E}{\bf \ul 92.2}      & \cellcolor[HTML]{FFFC9E}60.8                & \cellcolor[HTML]{FFFC9E}46.7                & \cellcolor[HTML]{FFFC9E}50.5          \\ \hline
  \end{tabular}
  \end{table}

  \subsection{Analysis of improvement}
  \label{seq:analysis}
  The API call improvement was much higher than that for other system response types (e.g., asking slots).
  Therefore, we analyzed the \apicall~results by focusing on \Match~(QA-LSTM and MMNs) and \ReRank.

  \Match: \Match~can use response candidates directly for the matching scores, whereas \BDS~simply uses the dialog history and the current user query, as shown in Fig~\ref{fig:arch}.
  Thus, \Match~has an advantage when some words are shared between the dialog history and the response candidate.
  \apicall~action contains slot entities that are mentioned by a user in the dialog history.
  That is why \Match~yields improvement in \apicall.
  Furthermore, \Match~can reflect the number of slots matching with the score, as apparent from Table \ref{tb:similar-cands}.
  Note that \Match~uses the cosine similarity, whereas \BDS~uses softmax to calculate scores.
  
  \begin{table}[ht]
      \small
      \caption{Sample comparison of top three candidates of \BDS~and \Match~for ASR-Task 6.
      The underscores indicate the correct slots. ``R\_{\it slot}'' means that the user does not care about the restaurant preference.
      }
\centering
\small
\label{tb:similar-cands}
\begin{tabular}{cl|l|l}
\multicolumn{1}{l}{}                                                &   & Score & Predicted Answer                             \\ \hline                                                                        
\multirow{3}{*}{BDS}                                                & 1 & \bf .847  &   {\small \bf api\_call \underline{R\_cuisine} \underline{east} \underline{expensive}} \\
                                                                    & 2 & .089  &   {\small api\_call \underline{R\_cuisine} \underline{south} \underline{expensive}}  \\
                                                                    & 3 & .028  &   {\small you are looking for a restaurant...} \\ \hline
\multirow{3}{*}{MMNs}                                               & 1 & \bf .841  &   {\small \bf api\_call \underline{R\_cuisine} \underline{east} \underline{expensive}} \\
                                                                    & 2 & .701  &   {\small api\_call \underline{R\_cuisine} \underline{east} R\_price}  \\
                                                                    & 3 & .695  &   {\small api\_call \underline{R\_cuisine} R\_location  \underline{expensive}} \\ \hline
\multirow{3}{*}{\begin{tabular}[c]{@{}c@{}}QA-\\ LSTM\end{tabular}} & 1 & \bf .774  &   {\small \bf api\_call \underline{R\_cuisine} \underline{east} \underline{expensive}} \\
                                                                    & 2 & .623  &   {\small api\_call british \underline{east} \underline{expensive}}  \\
                                                                    & 3 & .595  &   {\small api\_call british \underline{east}  R\_price} \\ \hline
\end{tabular}
\end{table}
    
    {\it RR1 (Rule-based)}:
    This \ReRank~strongly depends on the result of \Match~result, i.e., \(\alpha\) is only set to 1 for the top $N$ candidates in \Match~as Eq.~\ref{eq:rule-score} shows.
    Table~\ref{tb:result-n-best} indicates that \Match~has good accuracy results for the top K candidates compared to \BDS.
    This coverage contributes to the success of RR1.
    \begin{table}[ht]
      \small
      \centering
      \caption{Accuracy of top $K$ candidates. The ground truth label was included in the top $K$ predictions.}
          \begin{tabular}{r|rrr|rrr}
          \multicolumn{1}{l|}{} & \multicolumn{3}{c|}{bAbI/bAbI+}  & \multicolumn{3}{c}{ASR-Task 6}                                                  \\
        \multicolumn{1}{l|}{} & \multicolumn{1}{c}{BDS} & \multicolumn{1}{c}{MMNs} & \multicolumn{1}{c|}{\begin{tabular}[c]{@{}c@{}}QA-\\ LSTM\end{tabular}} & \multicolumn{1}{c}{BDS} & \multicolumn{1}{c}{MMNs} & \multicolumn{1}{c}{\begin{tabular}[c]{@{}c@{}}QA-\\ LSTM\end{tabular}} \\ \hline
          1   & 82.1 & 84.7 & \bf 92.0  & 42.5 & 43.7      & \bf 46.2 \\ 
          2   & 87.4 & 92.7 & \bf 97.1  & 51.1 & \bf 53.9  & 52.8 \\ 
          3   & 90.1 & 95.4 & \bf 98.3  & 54.6 & 57.6      & \bf 58.2 \\ \hline
          \end{tabular}
      \label{tb:result-n-best}
    \end{table}

    {\it RR2 (Stacking and logistic regression)}:
    As mentioned above, our word-based matching approach works well for re-ranking. It uses both the predictions and the word features as input.
    Table~\ref{tb:ablation} presents the results of an ablation study to check whether the strength of word features of \Match~ propagates to \ReRank.
    While general stacking uses predictions from base classifiers only,
    we use additional features, i.e., the dialog context word features and the candidate response.
    If we exclude either context or answer embeddings, 
    the accuracy of Total changes only slightly; however the API accuracy significantly decreases significantly.
    We presume that word-embedding features are effective for system actions that use slot entities, as 
    removing both word features causes the Total accuracy to decrease dramatically.
    \begin{table}[!ht]
      \centering
      \small
      \caption{Ablation study on ASR-Task 6, where ``ctx'' indicates \(e_{ctx}\) and ``ans'' indicates \(e_{ans}\) in Eq.~\ref{eq:stacking_input}.}
      \label{tb:ablation}
      \begin{tabular}{l|ll|ll}
                     & \multicolumn{2}{c|}{MMNs}     & \multicolumn{2}{c}{QA-LSTM} \\
                              & Total     & API       & Total    & API    \\ \hline
        Full                  & \bf 44.8  & \bf 43.2  & 46.7     & \bf 50.5   \\ \hline
        w/o context           & 43.9      & 37.0      & 46.7     & 47.2   \\
        w/o answer            & 44.6      & 39.0      & \bf 46.8 & 47.0   \\
        w/o context \& answer & 43.8      & 39.3      & 45.0     & 41.9   \\ \hline
        \end{tabular}
    \end{table}

   \section{Conclusions and Future Work}
   This paper describes a context-aware dialog response re-ranking system for task-oriented dialog systems based on two key re-ranking modules; \Match~and \ReRank.
   We have boosted the performance of an existing task-oriented dialog system.

   We evaluated five \Match~models, and two \ReRank~models on the real human--computer restaurant search dialogs with speech recognition errors.
   Our framework improved the existing dialog system by using neural-based \Match~models and both \ReRank~models.
 
 To our knowledge none of the previous studies presented a re-ranking response task that uses response ranking to validate response candidates. We have presented a simple and effective re-ranking module to reorder response candidates.
   
   Our research should be extended to natural dialogs including hesitations and corrections since applying word-level attention may be key to resolving disfluencies. We should also apply this technique to other types of dialog systems (e.g., chit-chat and question answering).
   
   
  \bibliographystyle{IEEEbib}
  \bibliography{strings,refs}
  
  \end{document}